\documentclass[runningheads]{llncs}

\usepackage{graphicx}
\usepackage{comment}
\usepackage{amsmath,amssymb}
\usepackage{color}

\usepackage{mathtools}
\usepackage{physics}
\usepackage{booktabs}
\usepackage{multirow}
\usepackage{pifont}

\usepackage{url}
\usepackage{hyperref}

\newcommand\blfootnote[1]{%
  \begingroup
  \renewcommand\thefootnote{}\footnote{#1}%
  \addtocounter{footnote}{-1}%
  \endgroup
}

\newif\ifreview
\reviewfalse

\ifreview
	\usepackage{lineno}

	\linenumbers
\fi

\begin{document}


\def\SubNumber{120}

\def\GCPRTrack{Regular Track}

\title{CAGAN: Text-To-Image Generation with Combined Attention Generative Adversarial Networks}

\ifreview
\else

	\author{Henning Schulze \qquad Dogucan Yaman \qquad Alexander Waibel}
	
	\authorrunning{H. Schulze et al.}
	
	\institute{Institute for Anthropomatics and Robotics, Karlsruhe Institute of Technology, Germany}
\fi

\maketitle              


\begin{abstract}


Generating images according to natural language descriptions is a challenging task. Prior research has mainly focused to enhance the quality of generation by investigating the use of spatial attention and/or textual attention thereby neglecting the relationship between channels. In this work, we propose the Combined Attention Generative Adversarial Network (CAGAN) to generate photo-realistic images according to textual descriptions. The proposed CAGAN utilises two attention models: word attention to draw different sub-regions conditioned on related words; and squeeze-and-excitation attention to capture non-linear interaction among channels. With spectral normalisation to stabilise training, our proposed CAGAN improves the state of the art on the IS and FID on the CUB dataset and the FID on the more challenging COCO dataset. Furthermore, we demonstrate that judging a model by a single evaluation metric can be misleading by developing an additional model adding local self-attention which scores a higher IS, outperforming the state of the art on the CUB dataset, but generates unrealistic images through feature repetition. 
\end{abstract}


\blfootnote{\noindent \normalsize 
The final published version of the paper can be found here: \url{https://link.springer.com/chapter/10.1007/978-3-030-92659-5_25} (DOI: 10.1007/978-3-030-92659-5\_25)
}
\section{Introduction}
\label{sec:introduction}

Generating images according to natural language descriptions spans a wide range of difficulty, from generating synthetic images to simple and highly complex real-world images. It has tremendous applications such as photo-editing, computer-aided design, and may be used to reduce the complexity of or even replace rendering engines~\cite{Ph2016}. Furthermore, good generative models involve learning new representations. These are useful for a variety of tasks, for example classification, clustering, or supporting transfer among tasks.

Although generating images highly related to the meanings embedded in a natural language description is a challenging task due to the gap between text and image modalities, there has been exciting recent progress in the field using numerous techniques and different inputs~\cite{LiPe2019}~\cite{Hi2019} ~\cite{LiZh2019}~\cite{Zhu2019}~\cite{Yi2019}~\cite{Qia2019}~\cite{Ca2019}~\cite{Li2019}~\cite{Ch2019}~\cite{LiWu2019} \cite{Xu2018}~\cite{Reed2016}~\cite{Re2016} yielding impressive results on limited domains. A majority of approaches are based on Generative Adversarial Networks (GANs)~\cite{Go2014}. Zhang et al. introduced Stacked GANs~\cite{Zh2017} which consist of two GANs generating images in a low-to-high resolution fashion. The second generator receives the image encoding of the first generator and the text embedding as input to correct defects and generate higher resolution images. Further research has mainly focused to enhance the quality of generation by investigating the use of spatial attention and/or textual attention thereby neglecting the relationship between channels.




\begin{figure}[t]
\centering
  \includegraphics[width=0.80\textwidth]{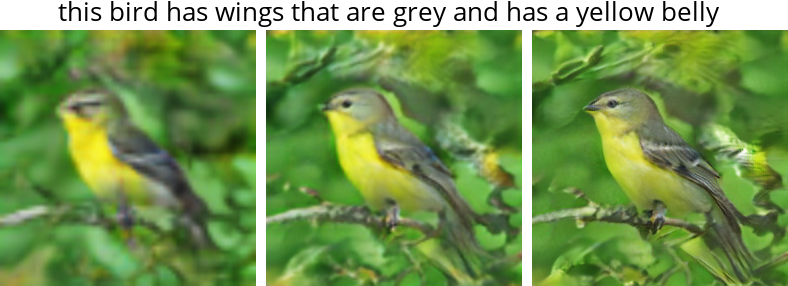}
  \caption{Example results of the proposed CAGAN (SE). The generated images are of $64$x$64$, $128$x$128$, and $256$x$256$ resolutions respectively, bilinearly upsampled for visualization.}
  \label{fig:Cover}
\end{figure}

In this work, we propose Combined Attention Generative Adversarial Network (CAGAN) that combines multiple attention models, thereby paying attention to word, channel, and spatial relationships. First, the network uses a deep bi-directional LSTM encoder~\cite{Xu2018} to obtain word and sentence features. Then, the images are generated in a coarse to fine fashion (see \autoref{fig:Cover}) by feeding the encoded text features into a three stage GAN. Thereby, we utilise local-self attention~\cite{Ra2019}  mainly during the first stage of generation; word attention at the beginning of the second and the third generator; and squeeze-and-excitation attention~\cite{Hu2018} throughout the second and the third generator. We use the publicly available CUB~\cite{Wa2011} and COCO~\cite{Li2014} datasets to conduct the experimental analysis. Our experiments show that our network generates images of similar quality as previous work while either advancing or competing with the state of the art on the Inception Score (IS)~\cite{Sa2016} and the Fréchet Inception Distance (FID)~\cite{He2017}.

The main contributions of this paper are threefold:

(1) We incorporate multiple attention models, thereby reacting to subtle differences in the textual input with fine-grained word attention; modelling long-range dependencies with local self-attention; and capturing non-linear interaction among channels with squeeze-and-excitation (SE) attention. SE attention can learn to learn to use global information to selectively emphasise informative features and suppress less useful ones.

(2) We stabilise the training with spectral normalisation~\cite{Mi2018}, which restricts the function space from which the discriminators are selected by bounding the Lipschitz norm and setting the spectral norm to a designated value.

(3) We demonstrate that improvements on single evaluation metrics have to be viewed carefully by showing that evaluation metrics may react oppositely.

The rest of the paper is organized as follows: In Section \ref{sec:relatedWork}, we give a brief overview of the literature. In Section \ref{sec:method}, we explain the presented approach in detail. In Section \ref{sec:experiments}, we mention the employed datasets and experimental results. Then, we discuss the outcomes and we conclude the paper in Section \ref{sec:conclusion}.

\section{Related Work}
\label{sec:relatedWork}

While there has been substantial work for years in the field of image-to-text translation, such as image caption generation~\cite{Ba2018}~\cite{Fe2015}~\cite{Xu2015}, only recently the inverse problem came into focus: text-to-image generation. Generative image models require a deep understanding of spatial, visual, and semantic world knowledge. A majority of recent approaches are based on GANs~\cite{Go2014}.

Reed et al.~\cite{Re2016} use a GAN with a direct text-to-image approach and have shown to generate images highly related to the text's meaning. Reed et al.~\cite{Reed2016} further developed this approach by conditioning the GAN additionally on object locations. Zhang et al. built on Reed et al.'s direct approach developing StackGAN~\cite{Zh2017} generating 256x256 photo-realistic images from detailed text descriptions. Although StackGAN yields remarkable results on specific domains, such as birds or flowers, it struggles when many objects and relationships are involved. Zhang et al.~\cite{Zha2017} improved StackGAN by arranging multiple generators and discriminators in a tree-like structure, allowing for more stable training behaviour by jointly approximating multiple distributions. Xu et al.~\cite{Xu2018} introduced a novel loss function and fine-grained word attention into the model.


Recently, a number of works built on Xu et al.'s~\cite{Xu2018} approach: Cheng et al.~\cite{Ch2019} employed spectral normalisation~\cite{Mi2018} and added global self-attention to the first generator; Qiao et al.~\cite{Qi2019} introduced a semantic text regeneration and alignment module thereby learning text-to-image generation by redescription; Li et al.~\cite{Li2019} added channel-wise attention to Xu et al.'s spatial word attention to generate shape-invariant images when changing text descriptions; Cai et al.~\cite{Ca2019} enhanced local details and global structures by attending to related features from relevant words and different visual regions; Yin et al.~\cite{Yi2019} focused on disentangling the semantic-related concepts and introduced a contrasive loss to strengthen the image-text correlation; and Zhu et al.~\cite{Zhu2019} refined Xu et al.'s fine-grained word attention by dynamically selecting important words based on the content of an initial image.

 
Instead of using multiple stages or multiple GANs, Li et al.~\cite{LiWu2019} used one generator and three independent discriminators to generate multi-scale images conditioned on text in an adversarial manner. Johnson et al.~\cite{Jo2018} introduced a GAN that receives a scene graph consisting of objects and their relationships as input and generates complex images with many recognizable objects. However, the images are not photo-realistic. Qiao et al.~\cite{Qia2019} introduced LeicaGAN which adopts text-visual co-embeddings to convey the visual information needed for image generation.



Other approaches are based on autoencoders~\cite{Sn2017}~\cite{Do2017}~\cite{Wu2017}, autoregressive models~\cite{Oord2016}~\cite{Re2017}~\cite{Gu2018}, or other techniques~\cite{Th2015}~\cite{Ku2015}~\cite{Kul2015}~\cite{Xi2016}.


We propose to expand the focus of attention to channel, word and spatial relationships instead of a subset of these thereby enhancing the quality of generation.


\section{The framework of Combined Attention Generative Adversarial Networks}
\label{sec:method}

\subsection{Combined Attention Generative Adversarial Networks}
The proposed CAGAN utilises three attention models: word attention to draw different sub-regions conditioned on related words, local self-attention to model long-range dependencies, and squeeze-and-excitation attention to capture non-linear interaction among channels. 

\begin{figure*}[t]
\centering
  \includegraphics[width=0.90\textwidth]{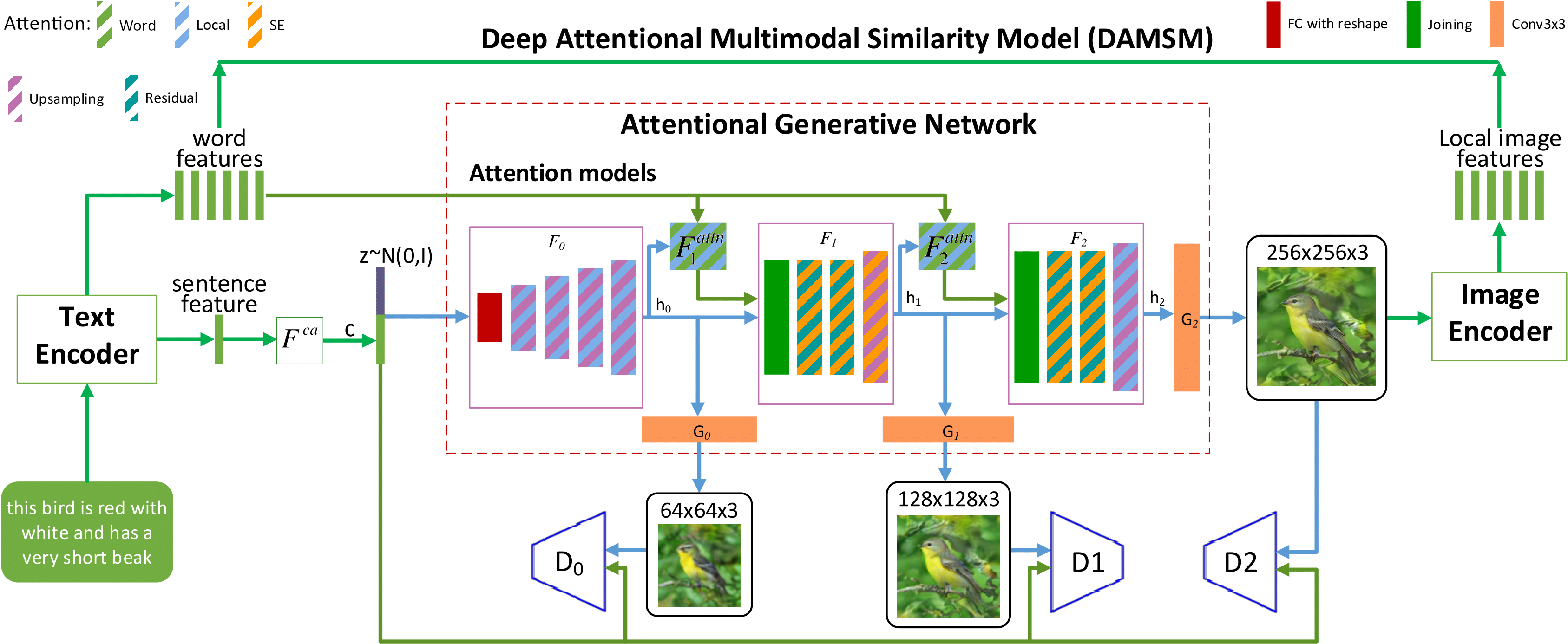}
  \caption{The architecture of the proposed CAGAN with word, SE, and local attention. When omitting local attention, local attention is removed from the $F_n^{attn}$ networks. In the upsampling blocks it is replaced by SE attention.}
  \label{fig:CAGAN}
\end{figure*}

The attentional generative model consists of three generators, which receive image feature vectors as input and generate images of small-to-large scales. First, a deep bidirectional LSTM encoder encodes the input sentence into a global sentence vector $s$ and a word matrix. Conditioning augmentation $F^{CA}$~\cite{Zh2017} converts the sentence vector into the conditioning vector. A first network receives the conditioning vector and noise, sampled from a standard normal distribution, as input and computes the first image feature vector. Each generator is a simple 3x3 convolutional layer that receives the image feature vector as input to compute an image. The remaining image feature vectors are computed by networks receiving the previous image feature vector and the result of the $i^{\text{th}}$ attentional model $F_i^{attn}$ (see \autoref{fig:CAGAN}), which uses the word matrix computed by the text encoder.

To compute word attention, the word vectors are converted into a common semantic space. For each subregion of the image a word-context vector is computed, dynamically representing word vectors that are relevant to the subregion of the image, i.e., indicating the weight the word attention model attends to the $l^{\text{th}}$ word when generating a subregion. The final objective function of the attentional generative network is defined as:
\begin{align}
L = L_G + \lambda L_{\text{DAMSM}}, \,\,\, where \,\, L_G = \sum_{i=0}^{m-1} L_{G_i} \,\, .
\label{eq:4.6}
\end{align}
Here, $\lambda$ is a hyperparameter to balance the two terms. The first term is the GAN loss that jointly approximates conditional and unconditional distributions~\cite{Zha2017}. At the $i^{\text{th}}$ stage, the generator $G_i$ has a corresponding discriminator $D_i$. The adversarial loss for $G_i$ is defined as:
\begin{small}
\begin{align}
L_{G_i} = \underbrace{-\frac{1}{2}\mathbb{E}_{\hat{y_i}\sim P_{G_i}} \big[ log(D_i(\hat{y_i})) \big] }_\text{unconditional loss} - \underbrace{\frac{1}{2}\mathbb{E}_{\hat{y_i}\sim P_{G_i}} \big[ log(D_i(\hat{y_i}, s)) \big] }_\text{conditional loss},
\label{eq:4.7}
\end{align}
\end{small}
where $\hat{y_i}$ are the generated images. The unconditional loss determines whether the image is real or fake while the conditional loss determines whether the image and the sentence match or not. Alternately to the training of $G_i$, each discriminator $D_i$ is trained to classify the input into the class of real or fake by minimizing the cross-entropy loss.



The second term of \autoref{eq:4.6}, $L_{\text{DAMSM}}$, is a fine-grained word-level image-text matching loss computed by the DAMSM~\cite{Xu2018}. The DAMSM learns two neural networks that map subregions of the image and words of the sentence to a common semantic space, thus measuring the image-text similarity at the word level to compute a fine-grained loss for image generation. The image encoder prior to the DAMSM is built upon a pretrained Inception-v3 model~\cite{Sz2016} with added perceptron layers to extract visual feature vectors for each subregion of the image and a global image vector.

\subsection{Attention models}
\subsubsection{Local self-attention} Similar to a convolution, local self-attention~\cite{Ra2019} extracts a local region of pixels $ab \in \mathcal{N}_k(i,j)$ for each pixel $x_{ij}$ and a given spatial extent $k$. An output pixel $y_{ij}$ computes as follows:	
\begin{align}
y_{ij} = \sum_{a,b \in \mathcal{N}_k(i,j)} \text{softmax}_{ab} (q_{ij}^T k_{ab}) v_{ab} \,\, .
\label{eq:4.11}
\end{align}
$q_{ij} = W_Q x_{ij}$ denotes the queries, $k_{ab} = W_K x_{ab}$ the keys, and $v_{ab} = W_V x_{ab}$ the values, each obtained via linear transformations $W$ of the pixel $ij$ and their neighbourhood pixels. The advantage over a simple convolution is that each pixel value is aggregated with a convex convolution of value vectors with mixing weights ($\text{softmax}_{ab}$) parametrised by content interactions.

\subsubsection{Squeeze-and-excitation (SE) attention} 


Instead of focusing on the spatial component of CNNs, SE attention aims to improve the channel component by explicitly modelling interdependencies among channels via channel-wise weighting. Thus, they can be interpreted as a light-weight self-attention function on channels.

First, a transformation, which is typically a convolution, outputs the feature map $U$. Because convolutions use local receptive fields, each entry of $U$ is unaware of contextual information outside its region. A corresponding SE-block addresses this issue by performing a feature recalibration.

A squeeze operation aggregates the feature maps of $U$ across the spatial dimension ($H \times W$) yielding a channel descriptor. The proposed squeeze operation is mean-pooling across the entire spatial dimension of each channel. The resulting channel descriptor serves as an embedding of the global distribution of channel-wise features.

A following excitation operation $F_{ex}$ aims to capture channel-wise dependencies, specifically non-linear interaction among channels and non-mutually exclusive relationships. The latter allows multiple channels to be emphasized. The excitation operation is a simple self-gating operation with a sigmoid activation function:
\begin{align}
F_{ex}(z, W) = \sigma (g (z, W)) = \sigma (W_2 \delta(W_1 z)) \,\, ,
\label{eq:SE}
\end{align}
where $\delta$ refers to the ReLU activation function, $W_1 \in \mathbb{R}^{\frac{C}{r} \times C}$, and $W_2 \in \mathbb{R}^{C \times \frac{C}{r}}$. To limit model complexity and increase generalisation, a bottleneck is formed around the gating mechanism: a Fully Connected (FC) layer reduces the dimensionality by a factor of $r$. A second FC layer restores the dimensionality after the gating operation. The authors recommend an $r$ of $16$ for a good balance between accuracy and complexity ($\sim 10\%$ parameter increase on ResNet-50~\cite{He2016}). Ideally, $r$ should be tuned for the intended architecture.

The excitation operation $F_{ex}$ computes per-channel modulation weights. These are applied to the feature maps $U$ performing an adaptive recalibration.   
\section{Experiments}
\label{sec:experiments}



\subsubsection{Dataset} We employed CUB~\cite{Wa2011} and COCO~\cite{Li2014} datasets for the experiments. The CUB dataset~\cite{Wa2011} consists of 8855 train and 2933 test images. To perform evaluation, one image per caption in the test set is computed since each image has ten captions. The COCO dataset~\cite{Li2014} with the 2014 split consists of 82783 train and 40504 test images. We randomly sample $30000$ captions from the test set for the evaluation.





\subsubsection{Evaluation metric} In this work, we utilized the Inception Score (IS)~\cite{Sa2016} and The Fréchet Inception Distance (FID)~\cite{He2017} to evaluate the performance of proposed method. The IS~\cite{Sa2016} is a quantitative metric to evaluate generated images. It measures two properties: highly classifiable and diverse with respect to class labels. Although the IS is the most widely used metric in text-to-image generation, it has several issues~\cite{Ro2017}~\cite{Sh2018}~\cite{Au2017} regarding the computation of the score itself and the usage of the score. According to the authors of~\cite{Sh2018} it: "fails to provide useful guidance when comparing models". 

The FID~\cite{He2017} views features as a continuous multivariate Gaussian and computes a distance in the feature space between the real data and the generated data. A lower FID implies a closer distance between the generated image distribution and the real image distribution. The FID is consistent with human judgment and more consistent to noise than the IS~\cite{He2017} although it has a slight bias~\cite{Lu2018}. 
Please note that there is some inconsistency in how the FID is calculated in prior work, originating from different pre-processing techniques that significantly impact the score. We use the official implementation\footnote{\url{https://github.com/bioinf-jku/TTUR}} of the FID. To ensure a consistent calculation of all of our evaluation metrics, we replace the generic Inception v3 network with the pre-trained Inception v3 network we used for computing the IS of the corresponding dataset. 
We re-calculate the FID scores of papers with an official model to provide a fair comparison.

\subsubsection{Implementation detail} We employ spectral normalisation~\cite{Mi2018}, a weight normalisation technique to stabilise the training of the discriminator, during training. To compute the semantic embedding for text descriptions, we employ a pre-trained bi-direction LSTM encoder by Xu et al.~\cite{Xu2018} with a dimension of $256$ for the word embedding. The sentence length was $18$ for the CUB dataset and $12$ for the COCO dataset. 

All networks are trained using the Adam optimiser~\cite{Kin2014} with a batch size of $20$, a learning rate of $0.0002$, and $\beta_1 = 0.5$ and $\beta_2 = 0.999$. We train for $600$ epochs on the CUB and for $200$ epochs on the COCO dataset. For the model utilising squeeze-and-excitation attention we use $r = 1$, and $\lambda = 0.1$ and $\lambda = 50.0$, respectively for the CUB and the COCO dataset. For the model utilising local self-attention as well we use $r = 4$, and $\lambda = 5.0$ and $\lambda = 50.0$.



\begin{table}[t]
\centering
\caption{Fréchet Inception Distance (FID) and Inception Score (IS) of state-of-the-art models and our two CAGAN models on the CUB and COCO dataset with a 256x256 image resolution. The unmarked scores are those reported in the original papers. Scores marked with $^\text{\ding{61}}$ were calculated with a pre-trained model provided by the respective authors. $\uparrow$ ($\downarrow$) means the higher (lower), the better.}
\begin{tabular}{c c c c c c}
\toprule
\multicolumn{1}{c}{} & \multicolumn{2}{c}{CUB dataset} & \multicolumn{1}{c}{}& \multicolumn{2}{c}{COCO dataset}\\

\multirow{-2}{*}{Model} & \multicolumn{1}{c}{IS$\uparrow$} & \multicolumn{1}{c}{FID$\downarrow$} & \multicolumn{1}{c}{}& \multicolumn{1}{c}{IS$\uparrow$} & \multicolumn{1}{c}{FID$\downarrow$}\\
\midrule
Real Data & $25.52 \pm .09$ & $ 0.00 $ & & $37.97 \pm .88$ & $0.00$  \\
\midrule

AttnGAN~\cite{Xu2018}& $4.36 \pm .04$ & $47.76^\text{\ding{61}}$  & 
& $25.89 \pm .47$  & $31.05^\text{\ding{61}}$ \\

PPAN~\cite{LiWu2019}& $4.38 \pm .05$ & -  & & - & -   \\

HAGAN~\cite{Ch2019}& $4.43 \pm .03$ & -  & & - & -  \\

MirrorGAN~\cite{Qi2019}& $4.56 \pm .05$ & -  & & $26.47 \pm .41$ & -  \\

ControlGAN~\cite{Li2019}& $4.58 \pm .09$ & $49.18^\text{\ding{61}}$ & & $24.06 \pm .60$  & -  \\

DualAttn-GAN~\cite{Ca2019} & $4.59 \pm .07$ & -  & & - & - \\

LeicaGAN~\cite{Qia2019}& $4.62 \pm .06$ & -  & & - & -   \\

SD-GAN~\cite{Yi2019}& $4.67 \pm .09$ & -  & 
& $35.69 \pm .50$  & -  \\

DM-GAN~\cite{Zhu2019} & $4.75 \pm .07$ & $43.20^\text{\ding{61}}$  & & $30.49 \pm .57$ & $22.84^\text{\ding{61}}$ \\

Obj-GAN~\cite{LiZh2019} & - & -  & & $30.29 \pm .33$ & - \\

OP-GAN~\cite{Hi2019} & - & -  & & $27.88 \pm .12$ &$23.29^\text{\ding{61}}$ \\

CPGAN~\cite{LiPe2019} & - & -  & & $\mathbf{52.73 \pm .61}$ & $49.92^\text{\ding{61}}$ \\

\midrule
CAGAN\_SE (ours) & $4.78 \pm .06$  & $\mathbf{42.98}$  & & $32.60 \pm .75$ & $\mathbf{19.88}$ \\

CAGAN\_L+SE (ours) & $\mathbf{4.96 \pm .05}$ & $61.06$  & &  $33.89 \pm .69$ & $27.40$  \\
\bottomrule
\end{tabular}
\label{tab:stateOfTheArtComparison}
\end{table}

\subsection{Results}

\subsubsection{Quantitative Results} 
\begin{figure}[t]
\centering
  \includegraphics[width=1.0\textwidth]{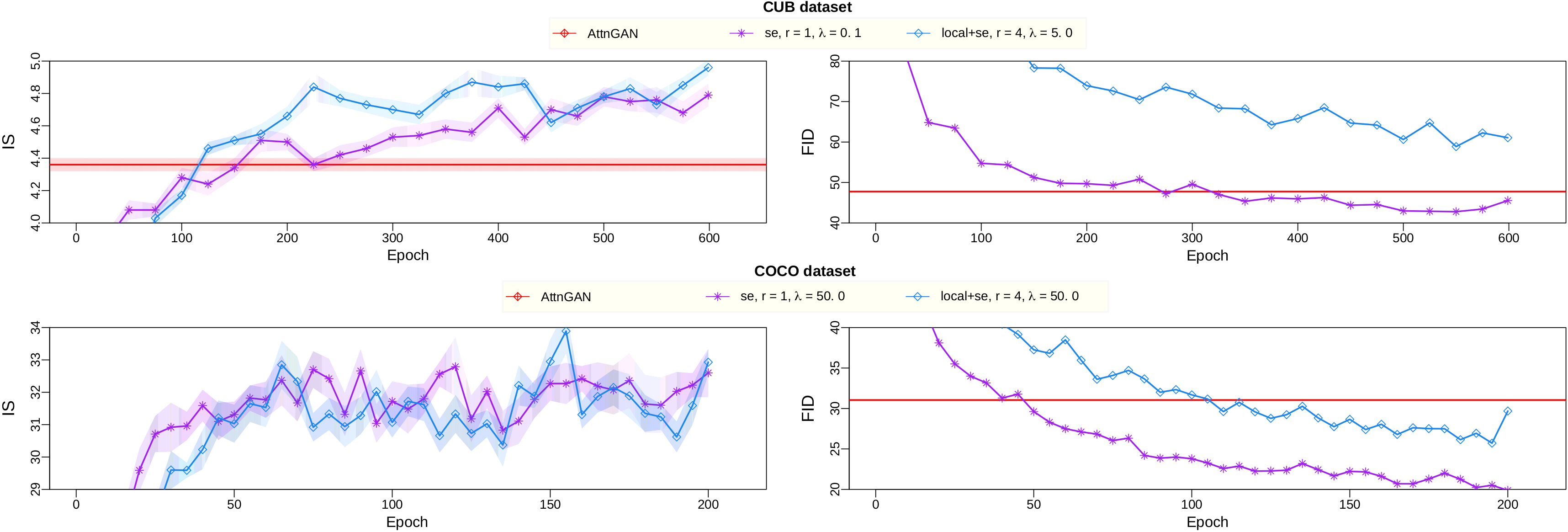}
  \caption{IS and FID of the AttnGAN~\cite{Xu2018}, our model utilising squeeze-and-excitation attention, and our model utilising squeeze-and-excitation attention and local self-attention on the CUB and the COCO dataset. The IS of the AttnGAN is the reported score and the FID was re-evaluated using the official model. The IS of the AttnGAN on the COCO dataset is with $25.89 \pm .47$ significantly lower than our models. We omitted the score to highlight the distinctions between our two models.}

  \label{fig:IS_FID}
\end{figure}

As \autoref{tab:stateOfTheArtComparison} and \autoref{fig:IS_FID} show, our model utilising squeeze-and-excitation attention outperforms the baseline AttnGAN~\cite{Xu2018} in both metrics on both datasets. The IS is improved by $9.6\% \pm 2.4\%$ and $25.9\% \pm 5.3\%$ and the FID by $10.0\%$ and $36.0\%$ on the CUB and the COCO dataset, respectively. Our approach also scores the highest IS and the highest FID on the CUB dataset and scores the best FID on the COCO dataset next to the third best IS.

\begin{figure}
\centering
  \includegraphics[width=1.0\textwidth]{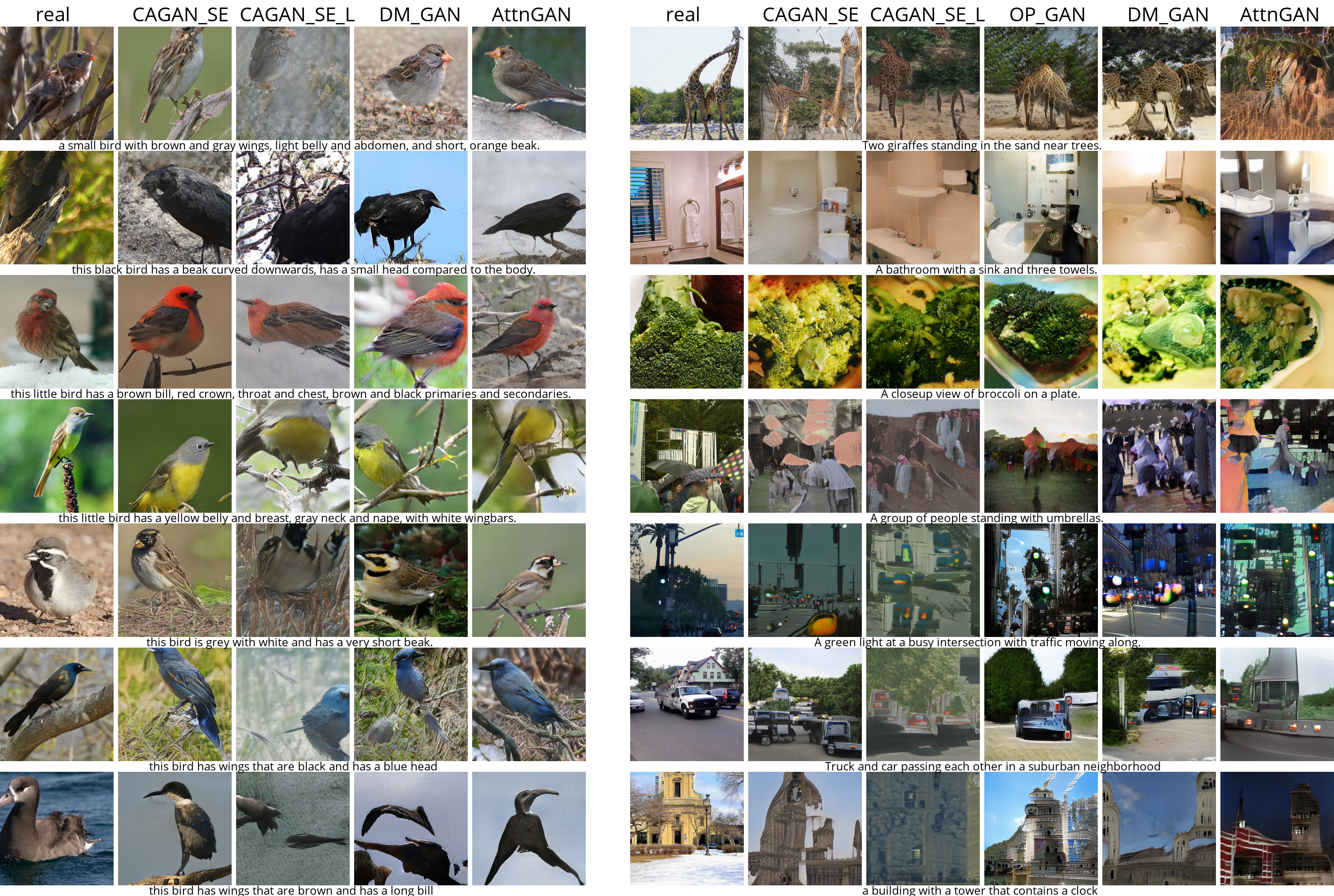}
  \caption{Comparison of images generated by our models (CAGAN\_SE and CAGAN\_SE\_L) with images generated by other current models~\cite{Hi2019}~\cite{Zhu2019}~\cite{Xu2018} on the CUB dataset (left) and on the more challenging COCO dataset (right).}

  \label{fig:Meta}
\end{figure}

Our second model, utilising squeeze-and-excitation attention and local self-attention, shows better IS scores than our other model. With $4.96 \pm 0.05$ it outperforms all other models on the CUB dataset, improving the state of the art by $4.4\% \pm 2.6\%$. However, it generates completely unrealistic images through feature repetitions (see \autoref{fig:Meta}) and has a major negative impact on the FID throughout training (see \autoref{fig:IS_FID}). This behaviour is similar to~\cite{LiPe2019} on the COCO dataset and demonstrates that a single score can be misleading and thus the importance of reporting both scores.

In summary, according to the experimental results, our proposed CAGAN achieved state-of-the-art results on both the CUB dataset and COCO dataset based on the FID metric. Moreover, we obtained state-of-the-art IS on the CUB dataset and quite a comparative IS on the COCO dataset. All these results indicate how our CAGAN model is effective for the text-to-image generation task.

\begin{figure}
\centering
  \includegraphics[page=1, width=1.0\textwidth]{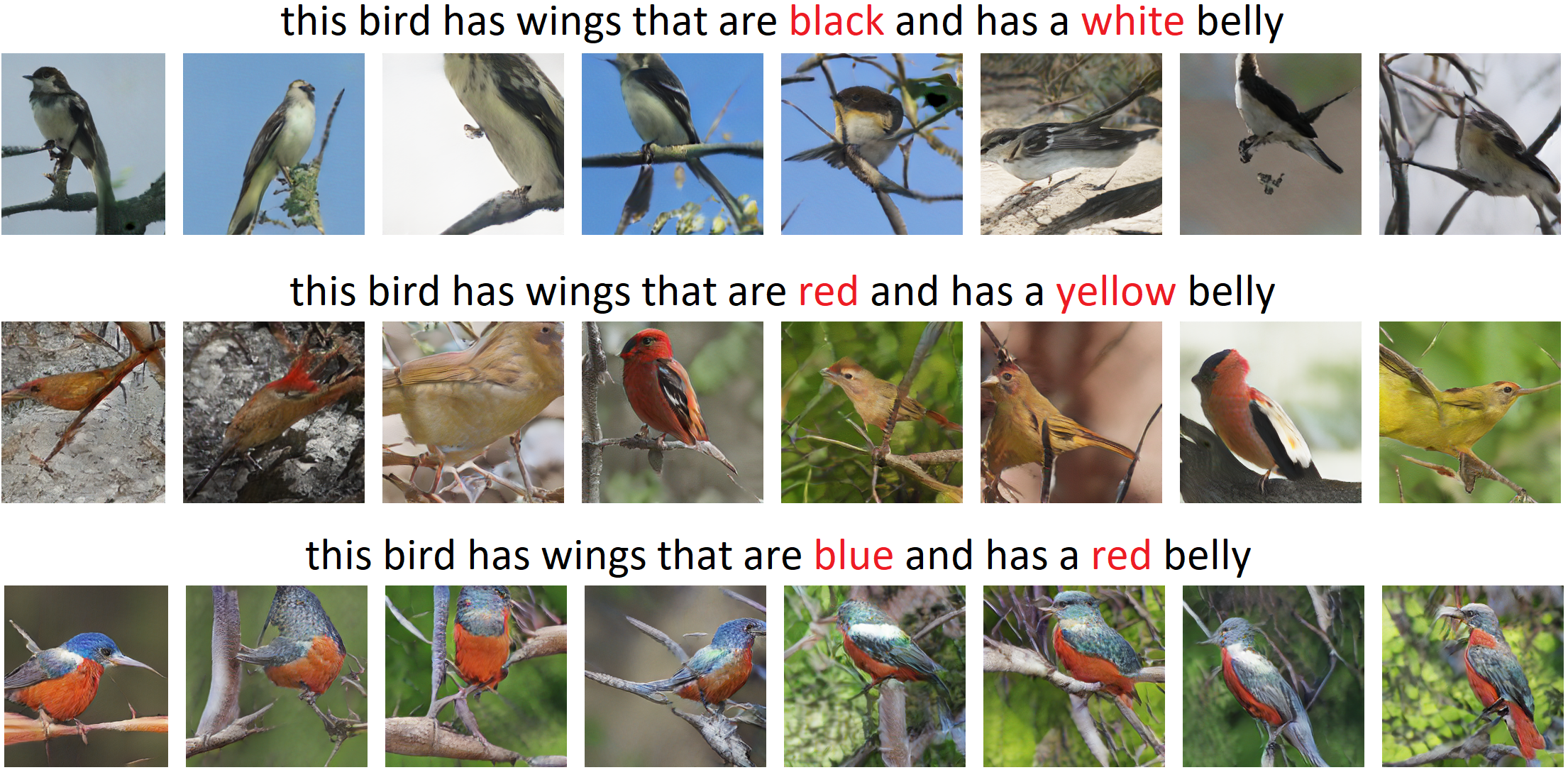}
  \caption{Example results of our SE attention model with $r = 1, \lambda = 0.1$ trained on the CUB dataset while changing some most attended, in the sense of word attention, words in the text descriptions.}

  \label{fig:SeTextAttention}
\end{figure}


\subsubsection{Qualitative Results}: \autoref{fig:Meta} shows images generated by our models and by several other models~\cite{Hi2019}~\cite{Zhu2019}~\cite{Xu2018} on the CUB dataset and on the more challenging COCO dataset. On the CUB dataset, our model utilising SE attention generates images of vivid details (see $1^{st}$, $4^{th}$, $5^{th}$, and $6^{th}$ row), demonstrating a strong text-image correlation (see $3^{th}$, $4^{th}$, and $5^{th}$ row), avoiding feature repetitions (see double beak, DM-GAN $6th$ row), and managing the difficult scene (see $7^{th}$ row) best. Cut-off artefacts occur in all presented models.

Our model incorporating local self-attention fails to produce realistic looking image, despite scoring higher ISs than the AttnGAN and our model utilising SE attention. Instead, it draws repetitive features manifesting in the form of multiple birds, drawn out birds, multiple heads, or strange patterns. The drawn features mostly match the textual descriptions. This provides a possible explanation why the model has a high IS despite scoring poorly on the FID: the IS cares mainly about the images being highly classifiable and diverse. Thereby, it presumes that highly classifiable images are of high quality. Our network demonstrates that high classify-ability and diversity and therefore a high IS can be achieved through completely unrealistic, repetitive features of the correct bird class. This is further evidence that improvements solely based on the IS have to be viewed sceptically.


On the more challenging COCO dataset, our model utilising SE attention demonstrates semantic understanding by drawing features that resemble the object, for example, the brown-white pattern of a giraffe ($1^{st}$ row), umbrellas ($4^{th}$ row), and traffic lights ($5^{th}$ row). Furthermore, our model draws distinct shapes for the bathroom ($2^{nd}$ row), broccoli ($3^{rd}$ row), and is the only one that properly approximates a tower building with a clock ($7^{th}$ row). Generally speaking, the results on the COCO dataset are not as realistic and robust as on the CUB dataset. We attribute this to the more complex scenes coupled with more abstract descriptions that focus rather on the category of objects than detailed descriptions. In addition, although there are a large number of categories, each category only has comparatively few examples thereby further increasing the difficulty for text-to-image-generation.

For our SE attention model we further test its generalisation ability by testing how sensitive the outputs are to changes in the most attended, in the sense of word attention, words in the text descriptions (see \autoref{fig:SeTextAttention}). The test is similar to the one performed on the AttnGAN~\cite{Xu2018}. The results illustrate that adding SE attention and spectral normalisation do not harm the generalisation ability of the network: the images are altered according to the changes in the input sentences, showing that the network retains its ability to react to subtle semantic differences in the text descriptions. 
\section{Conclusion}
\label{sec:conclusion}

In this paper, we propose the Combined Attention Generative Adversarial Network (CAGAN) to generate photo-realistic images according to textual descriptions. We utilise attention models such as, word attention to draw different sub-regions conditioned on related words; squeeze-and-excitation attention to capture non-linear interaction among channels; and local self-attention to model long-range dependencies. With spectral normalisation to stabilise training, our proposed CAGAN improves the state of the art on the IS and FID on the CUB dataset and the FID on the more challenging COCO dataset. Furthermore, we demonstrate that judging a model by a single evaluation metric can be misleading by developing an additional model which scores a higher IS, outperforming the state of the art on the CUB dataset, but generates unrealistic images through feature repetition. 



%
%
%
\bibliographystyle{splncs04}
\bibliography{egbib}

\end{document}